\documentclass[conference]{IEEEtran}

\usepackage{graphicx}
\usepackage{cite}
\usepackage{balance}
\usepackage[dvipsnames]{xcolor}
\usepackage{subcaption}  % For subfigures
\usepackage{booktabs}  % For better table formatting
\usepackage{array}     % For alignment
\usepackage[
    colorlinks=true,       % Use colored links
    linkcolor=blue,       % Color for internal links
    citecolor=blue,       % Color for citations
    urlcolor=blue,        % Color for URLs
    pdfborder={0 0 0}     % No border around links
]{hyperref}

\begin{document}

\title{Application of Multimodal Large Language Models in Autonomous Driving
}

\author{
  Md Robiul Islam \\
  Department of Computer Science \\
  William \& Mary \\
  \texttt{miislam@wm.edu}
  % \and
  % Sidi Lu\\
  % Department of Computer Science \\
  % William \& Mary \\
  % \texttt{sidi@wm.edu}
}

% make the title area
\maketitle

\begin{abstract}
In this era of technological advancements, several cutting-edge techniques are being implemented to enhance Autonomous Driving (AD) systems, focusing on improving safety, efficiency, and adaptability in complex driving environments. However, AD still faces some problems including performance limitations. To address this problem, we conducted an in-depth study on implementing the  Multi-modal Large Language Model. We constructed a Virtual Question Answering (VQA) dataset to fine-tune the model and address problems with the poor performance of MLLM on AD. We then break down the AD decision-making process into scene understanding, prediction, and decision-making. Chain of Thought has been used to make the decision more perfectly. Our experiments and detailed analysis of Autonomous Driving give an idea of how important MLLM is for AD. 
\end{abstract}

\begin{IEEEkeywords}
Autonomous Driving, MLLM, CoT, VQA, Fine-tuned, CogVLM2
\end{IEEEkeywords}
\section{Introduction}
Large Language Models (LLMs) have recently garnered significant attention for their exceptional ability to replicate human-like intelligence. These advancements have fueled growing excitement around Multimodal Large Language Models (MLLMs) \cite{yin2024survey}, which combine the advanced reasoning capabilities of LLMs with data from images, videos, and audio. This modality alignment empowers MLLMs to perform a wide range of tasks more efficiently, such as image classification, text-to-video matching, and speech recognition. Autonomous vehicles (AVs) have gained popularity due to advancements in computing technologies, as well as the introduction of Tesla's Autopilot \cite{ingle2016tesla}, Google's Waymo \cite{ettinger2021large}, and Baidu's Apollo \cite{feng2022application}. Existing deep learning-based solutions for analyzing, predicting, and making decisions for AVs rely significantly on available data. Deep learning-based AV systems frequently fail to respond effectively in corner instances, such as irregular road user behavior, unanticipated barriers, unfavorable weather conditions, and complicated traffic accidents ~\cite{zhou2022dynamically,huang2006autonomous,hu2023planning,chen2022milestones}. 
The current mainstream of the autonomous vehicle (AV) software pipeline includes several crucial modules, organized in a stack manner. Initially, the perception module consumes sensor data from Lidar, radar, and camera to provide object detection outcomes. Subsequently, the prediction module anticipates other relevant entities' high-level intentions and low-level trajectories based on scene analysis \cite{jiao2023chatgpt}. The planning module then makes decisions on the vehicle’s behavior, such as lane-keeping, gap tracking, and lane changing, followed by the 

\begin{figure}[ht]
 \centering
    % First image with subcaption
    \begin{minipage}[b]{0.48\textwidth}
        \centering
        \includegraphics[width=\textwidth]{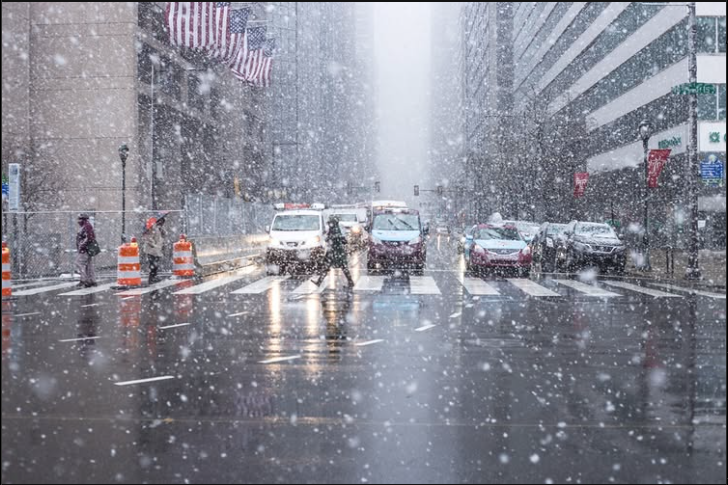}
        \subcaption{} % Subcaption for the first image
    \end{minipage}
\begin{minipage}[b]{0.45\linewidth}
 \includegraphics[width=1\textwidth]{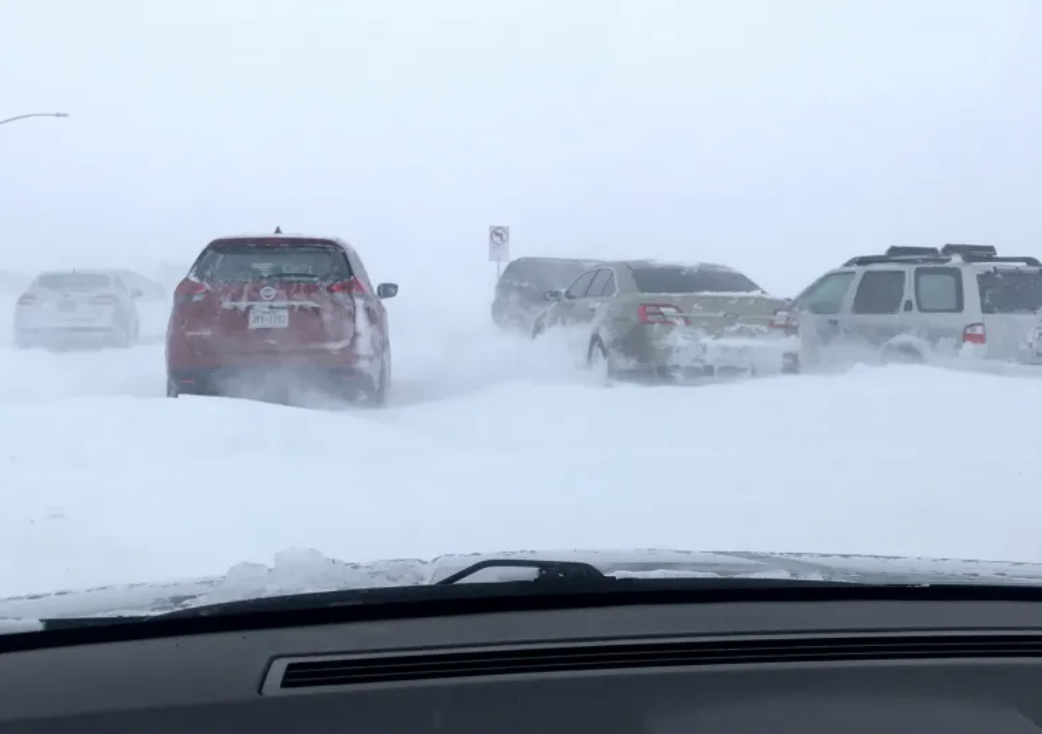}
 \subcaption{}
\end{minipage}
\begin{minipage}[b]{0.49\linewidth}
 \includegraphics[width=1\textwidth]{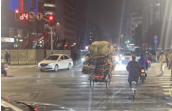}
  \subcaption{}
\end{minipage}%
\caption{(a) Road surfaces with water, which may result in improper AV positioning concerning the image of the water surface, are among the scenarios where AVs cannot effectively evaluate, forecast, and make judgments. (b) Roads covered in snow, where AVs find it difficult to make informed decisions because of a lack of information. (c) The intersection of construction where too complicated and unknown scenarios could cause AVs to make inaccurate predictions \cite{chen2024advanced}.}
\end{figure}

computation of trajectory-level waypoints for precise control tracking \cite{yang2023baichuan}.

Control inputs are applied to the ego vehicle, evolving it to transition to a new state following its physical dynamics and constraints. Deep neural networks (DNNs) have become integral to perception and prediction within the AV pipeline, with a growing interest in their application in planning and control. However, the black-box nature of DNNs, along with their inherent uncertainties from learning algorithms, presents challenges in ensuring the safety of closed-loop AV systems. These challenges are exacerbated by the generalizability issue faced by DNNs and the prevalence of long-tail driving scenarios not covered during training \cite{fu2022complexity}. Because of these, there is a lack of public confidence in AV due to the uncertain safety issues and most importantly the fact that the driving behaviors made by AV are uninterpretable without reasoning to the user.
As the capabilities of autonomous vehicles have expanded, so too have the challenges associated with ensuring their safety and reliability. Traffic safety has always been a critical concern in the development of autonomous vehicles, as these systems must be able to respond to a wide range of dynamic and unpredictable situations on the road. This is particularly important in the context of safety-critical events, such as sudden changes in traffic patterns, unexpected obstacles, and potential collisions. Additionally, traffic safety is influenced not only by the type of vehicle—whether conventional or autonomous—but also by the volume of traffic, as higher traffic volumes increase the probability of unsafe conditions\cite{abu2024using}.

This study examines the use of MLLMs as the primary decision-making tool for Autonomous Driving (AD), especially in corner instances. We examined the offline deployment of MLLMs in AVs. We suggested a paradigm for integrating MLLMs into a computational platform similar to an autonomous vehicle, resulting in an AD Agent. We developed a Visual Question-Answer (VQA) dataset and refined the CogVLM2 instead of CogVLM \cite{chen2024advanced} model to focus on AD tasks. To reduce MLLM illusions and improve interpretability, we developed a step-by-step Chain-of-Thought (CoT). The CoT splits the decision-making process into three phases: scene understanding, prediction, and choice. Each phase undergoes rigorous experimental validation. This approach enabled us to thoroughly evaluate the potential value and the applicability of MLLMs in many domains of AD.

Specifically, our contributions are as follows:

\begin{itemize}
    \item Our MLLM-based framework effectively executes AD tasks with limited computer resources, few shots, multi-modality, and complex scenarios. This attempt provides new insights and opportunities for the future deployment of more adaptable audiovisual systems
    \item We optimized generic MLLMs as AV Agents by creating a VQA dataset and establishing a reasoning chain. This reduced model illusions and improved attention.
    \item We validated MLLMs for scene interpretation, analysis, and decision-making using a real-world dataset. Experiments utilizing highwayenv \cite{highway-env} showed that MLLM-driven AV systems provide significant performance improvements. We discussed future directions and suggested techniques for improvement.

\end{itemize}

The rest of the paper is organized as follows: Section II discusses the background and the related work on AD. The experimental details are discussed in Section III. Experimental results are shown in Section IV. The discussion is in Section V and the Conclusion is in Section VI.
\section{Background \& Related Work}
\subsection{Traditional Autonomous Driving}
The Society of Automotive Engineers \cite{sae} defines driving automation as Levels 0 (no automation) to 5 (complete automation). enhancing autonomy reduces the need for human involvement while enhancing the vehicle's ability to understand its surroundings. AD solutions fall into two categories: conventional modular paradigms and end-to-end approaches \cite{codevilla2018end}. The modular approach divides the AV task into subtasks, which are executed in distinct modules.
While this approach has advantages such as flexibility and functional generality, it also presents issues in optimizing the pipeline and managing error propagation. According to UniAD \cite{hu2023planning}, the process can be divided into three stages: perception, prediction, and planning, with a focus on planning. These methods are typically easy to create. However, their lack of interpretability makes it difficult to diagnose issues, ensure safety, and comply with traffic rules. Automated systems may still fail in some driving scenarios, such as strong weather, bad lighting, or uncommon events \cite{10.1016/j.adhoc.2018.12.006}.

\subsection{Modular Autonomous Driving}
The modular autonomous driving system is a frequently used architecture in this arena \cite{yurtsever2020survey}. The autonomous driving task is divided into four separate components: perception, prediction, planning, and control. Each module is developed separately is in charge of specialized functionality throughout the system.
The planning module determines the best way from the present position to the destination depending on the vehicle's state and environmental conditions.
The process is typically separated into global and local planning. Global planning optimizes routes from a starting point to a destination using methods such as A* \cite{hart1968formal} and Dijkstra algorithms to search on a map. Local planning entails making real-time adjustments based on the vehicle's present circumstances, focusing on immediacy and reliability. Common approaches for local planning include RPP \cite{barraquand1991robot}, RRT \cite{lavalle2001randomized}, RRT* \cite{karaman2011sampling}, and others. Deep learning-based planners ~\cite{teng2023motion,zhu2022multi,gao2022cola,chen2018parallel} have become effective alternatives to classic approaches in recent years.

\subsection{Large Language Models}
Humans communicate and transmit information mostly through natural language. Models of natural language that are intended to understand and process natural language have changed during years of growth. Transformer architecture's highly parallelized data processing mechanism and potent performance \cite{vaswani2017attention} caused a disruptive revolution in the field of natural language processing (NLP). Another significant development in natural language models was BERT \cite{devlin2018bert}, which suggested pre-training the model on enormous volumes of unlabeled corpus data before fine-tuning it on particular tasks. It greatly improves baseline performance on a variety of NLP tasks. Language models having an enormous number of parameters—typically one billion or more—are referred to as large language models or LLMs. A recent work \cite{kaplan2020scaling} establishes the scaling law and demonstrates how language model performance is influenced by the number of model parameters, dataset size, and training computation. The most distinctive feature of LLMs is their display of emergent capabilities, which are typically absent from smaller models and include the ability to follow instructions, strong multi-step reasoning skills, and few-shot or zero-shot transfer learning across multiple downstream tasks. ChatGPT, notably GPT-3.5 \cite{brown2020language} and GPT-4 [\cite{achiam2023gpt}, is a key milestone in the development of LLMs. GPT-3.5's outstanding performance has garnered attention since its initial release. Researchers are leveraging LLMs' verbal knowledge, interpretation, analysis, and reasoning to address previously unsolvable issues. Open-source LLMs like as Llama2 \cite{touvron2023llama}, Qwen \cite{bai2023qwen}, and Phi \cite{li2023textbooks} are gaining attention from academia and industry, with some achieving equivalent performance to ChatGPT in specific tasks.

\begin{figure*}
    \centering
    \includegraphics[width=1\linewidth]{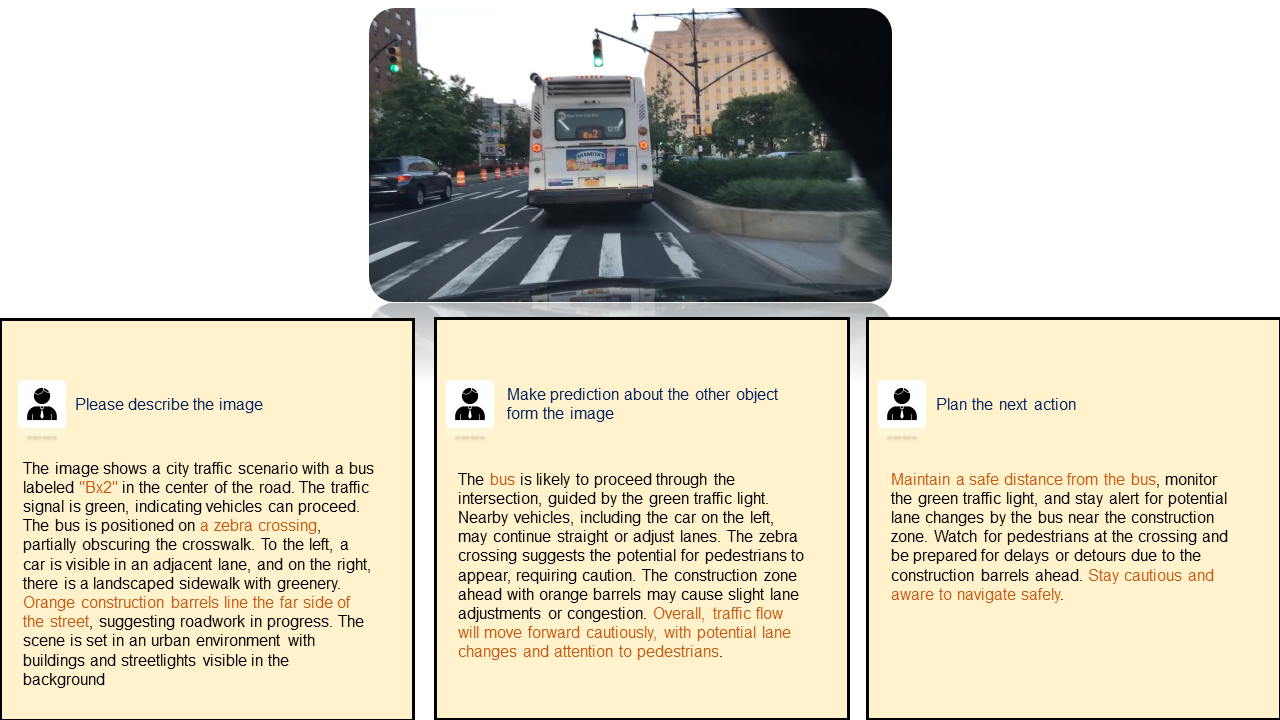}
    \caption{The model's step-by-step thinking chain generates information in a progressive manner, leading to more interpretable results.}
    \label{fig:enter-label}
\end{figure*}

\subsection{Vision Language Models}

By bridging the gaps between textual and visual information, Vision-Language Models (VLMs) link multimodal data by combining the powers of Natural Language Processing (NLP) and Computer Vision (CV). VLMs can understand the intricate link between natural languages and visual material by learning cross-modality data. Recently, as LLMs have grown in popularity, more attention has been paid to investigating the best ways to integrate visual modules into LLMs so they can carry out multimodal tasks.
\section{Methodology}
The overall methodology will be discuss in this section.

\subsection{Visual Question Answering (VQA) dataset build}
MLLMs may perform few- or even zero-shot learning tasks due to their extensive prior knowledge. Traditional open-source strategies prioritize versatility above specific domain competencies to improve generalizability. We fine-tuned the model to target AD tasks better and address the illusion problem. To address the scarcity of relevant datasets, we created a VQA dataset using common instances from BDD100k \cite{yu2020bdd100k} and Kitti \cite{Geiger2012CVPR}. The user prompt and the response is in English instead of other languages such as Bangla \cite{sentiment}.

\begin{figure}
    \centering
    \includegraphics[width=0.9\linewidth]{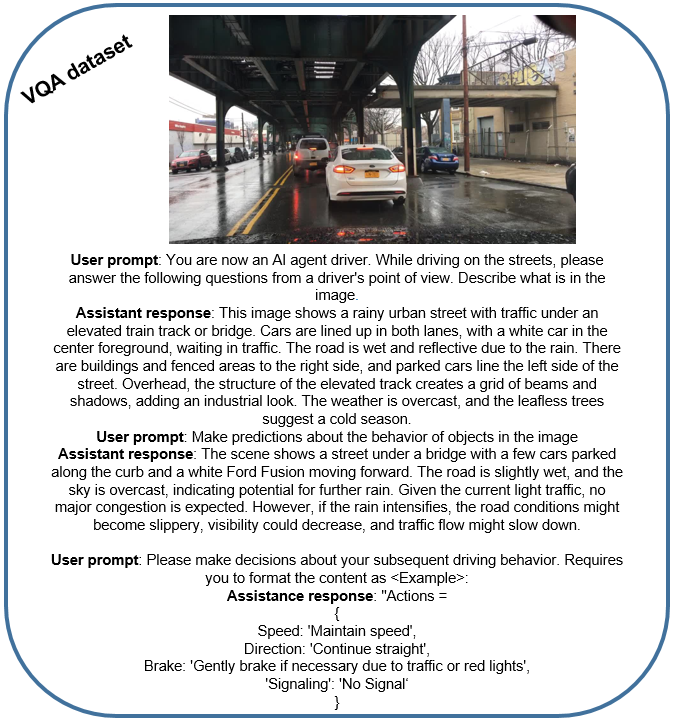}
    \caption{Visual Question Answer (VQA) dataset}
    \label{fig:vqa}
\end{figure}

In the beginning, we chose 100 photos for hand annotation, motivated by Segment Anything \cite{kirillov2023segment}. Using context-based learning, we then used ChatGPT-4 and Gemini to automatically annotate more photographs. After that, we carefully reviewed and improved the annotated material, which led to any samples that didn't satisfy requirements should be re-annotated using ChatGPT-4 and Gemini. We annotated 100 examples in a VQA dataset following four iterations. The dataset is mainly divided into three sections, as shown in Fig. \ref{fig:vqa}: Visual, Question, and Answer. For the model to function more expertly and comprehend the task process better, the quality of the dataset construction is essential. We developed a Chain of Thought (COT) that divides the AD task into smaller tasks that must be completed one at a time using three-part question-and-answer data instances. To ensure the model's output can be appropriately decoded by the action decoder, we adapted the output format in the third part of the Answer to fit its criteria. 

\begin{figure*}
    \centering
    \includegraphics[width=\linewidth]{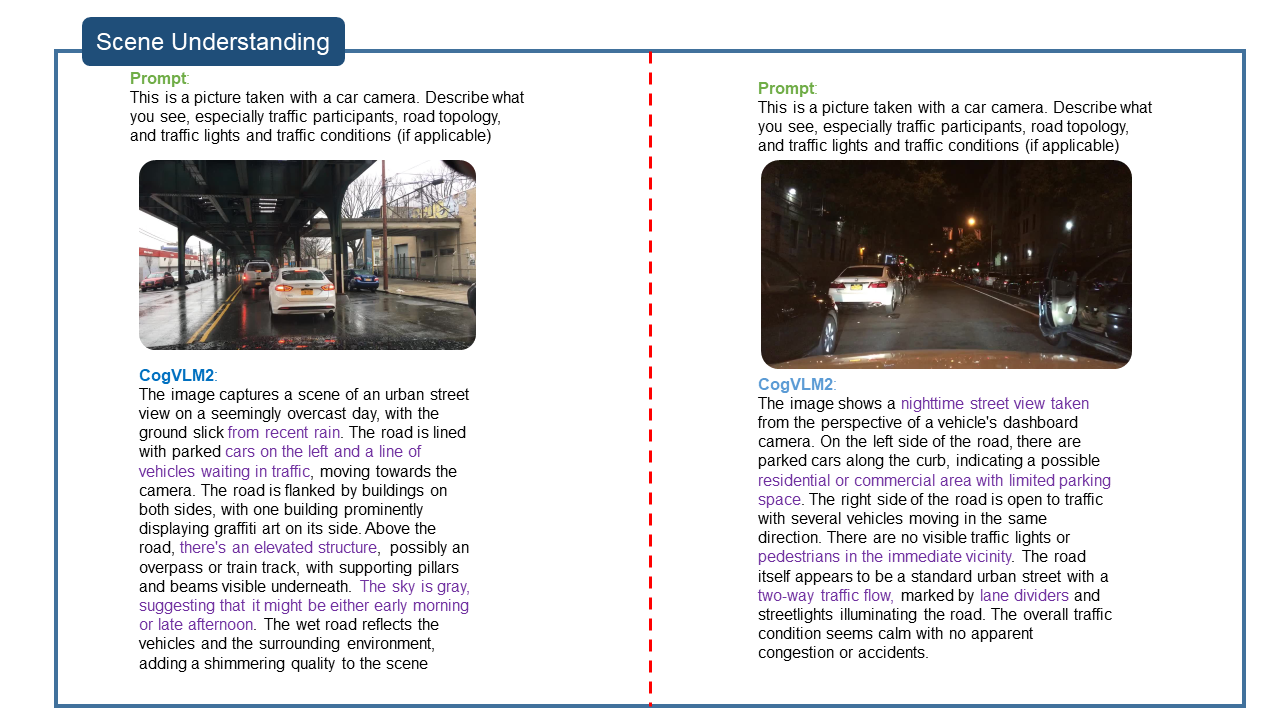}
    \caption{Used the fine-tuned CogVLM2 to describe traffic environments. The \textbf{purple} color text emphasizing the answer. The model can easily understand the recent rain, the elevated structure etc. }
    \label{fig:scene-un}
\end{figure*}

\section{Experimental Design}

\subsection{Hardware Setup}
\begin{figure}[!ht]
    \centering
    \begin{subfigure}[b]{0.15\textwidth}
        \centering
        \includegraphics[height=1.2in]{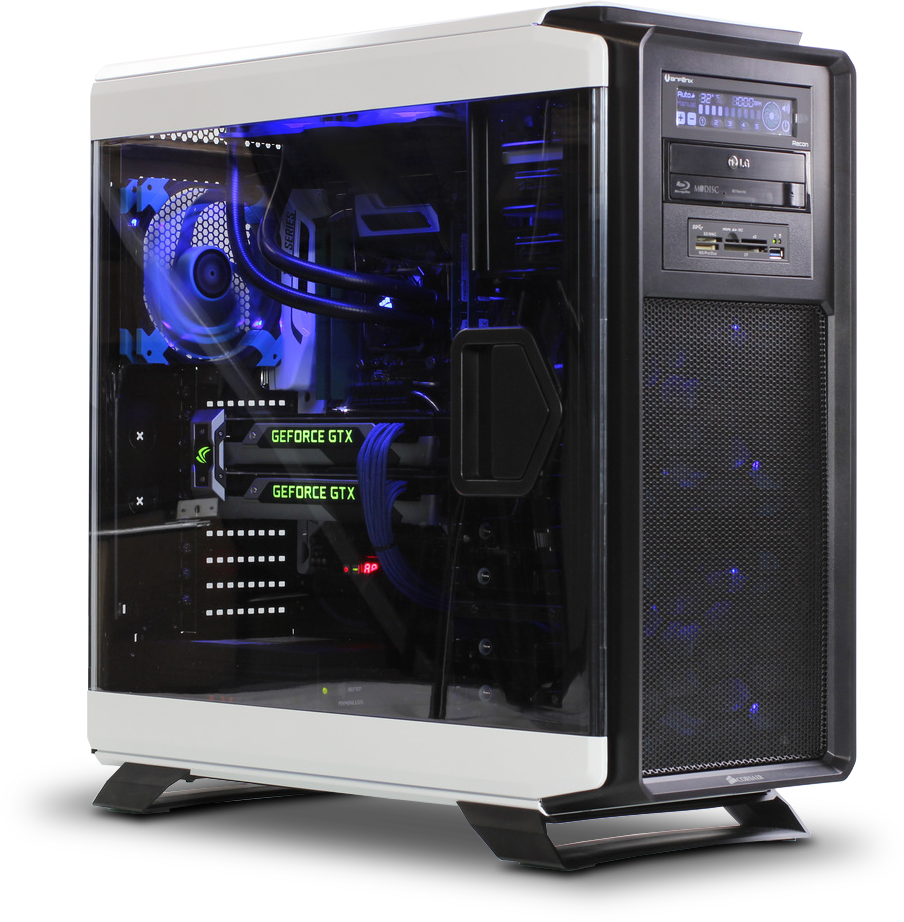}
    \end{subfigure}%
   \hfill
    \begin{subfigure}[b]{0.3\textwidth}
        \centering
        \includegraphics[height=1.2in]{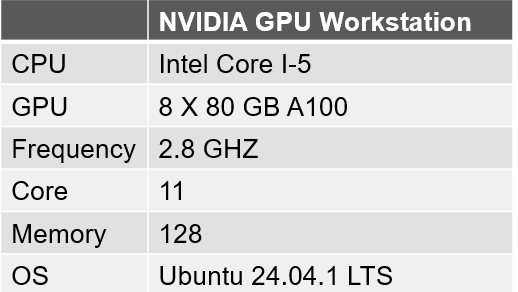}
    \end{subfigure}
    \caption{The configuration of NVIDIA GPU workstation}
    \label{fig:nvidia}
\end{figure}

Our hardware arrangement includes an NVIDIA GPU Workstation, made for high-demand workloads, and detailed information is presented in the accompanying Fig. \ref{fig:nvidia}. It features an Intel Core-i5 CPU, which is ideal for tackling complicated jobs effectively. This CPU excels at handling numerous tasks at once. The workstation also has a lot of RAM, which makes it easier to multitask and manage large datasets. This is useful for data analysis, machine learning, and running multiple programs or simulations simultaneously. This solution relies heavily on eight NVIDIA A100 graphics cards, each with 80 GB of memory. This arrangement has an impressive parallel processing capability, which is essential for demanding activities such as deep learning, rendering, and complex graphical computations. The NVIDIA A100 is known for its remarkable performance in professional applications, especially in Large Language Models.

\subsection{Why CogVLM2 outperform CogVLM}

In this paper, we use the updated version of CogVLM. In the paper \cite{chen2024advanced}, they used CogVLM. There are several differences between CogVLM and CogVLM2. CogVLM2 outperforms CogVLM in various aspects such as larger parameter size. Also, CogVLM2 was trained with a more diverse dataset. Here is the key difference between CogVLM and CogVLM2. \newline
\textbf{Performance Enhancement}:
CogVLM2 outperforms its predecessor in various benchmarks, notably achieving higher accuracy in tasks like TextVQA, DocVQA, and ChartQA. For instance, it scores higher in text-based visual question answering and document-based tasks, which demonstrates its improved comprehension and reasoning capabilities for images and text.
\newline
\textbf{Expanded Language Support}:
While CogVLM primarily focused on English, CogVLM2 offers enhanced support for both English and Chinese, broadening its applicability across different language contexts.
\newline
\textbf{Extented context length}:
CogVLM2 supports an extended text context length of up to 8,000 tokens and can handle images with resolutions up to 1344x1344 pixels, allowing it to process more detailed images and longer texts effectively.
\newline
\textbf{Base model and architecture}:
LLaMA 3, provides a more robust backbone model with improved foundational performance, making it more competitive with other state-of-the-art visual language models.

\subsection{Scene Understanding based on Multi-modal LLM}
We conducted a series of tests to assess the ability of Multi-Modal Language Models (MLLMs) to process and comprehend multimodal input. Fig. \ref{fig:scene-un} shows a significant example from these studies, demonstrating the current model's capacity to thoroughly understand visual input. In this figure, we show CogVLM2's scene analysis and understanding using road photos taken by onboard cameras. The model performed well in numerous scenarios, identifying single and dual carriageways, appropriately rating traffic conditions, and correctly detecting cars and people in the photos.

\subsection{Prediction based on Multi-modal LLM}
\begin{figure*}[!ht]
    \centering
    \includegraphics[width=\linewidth]{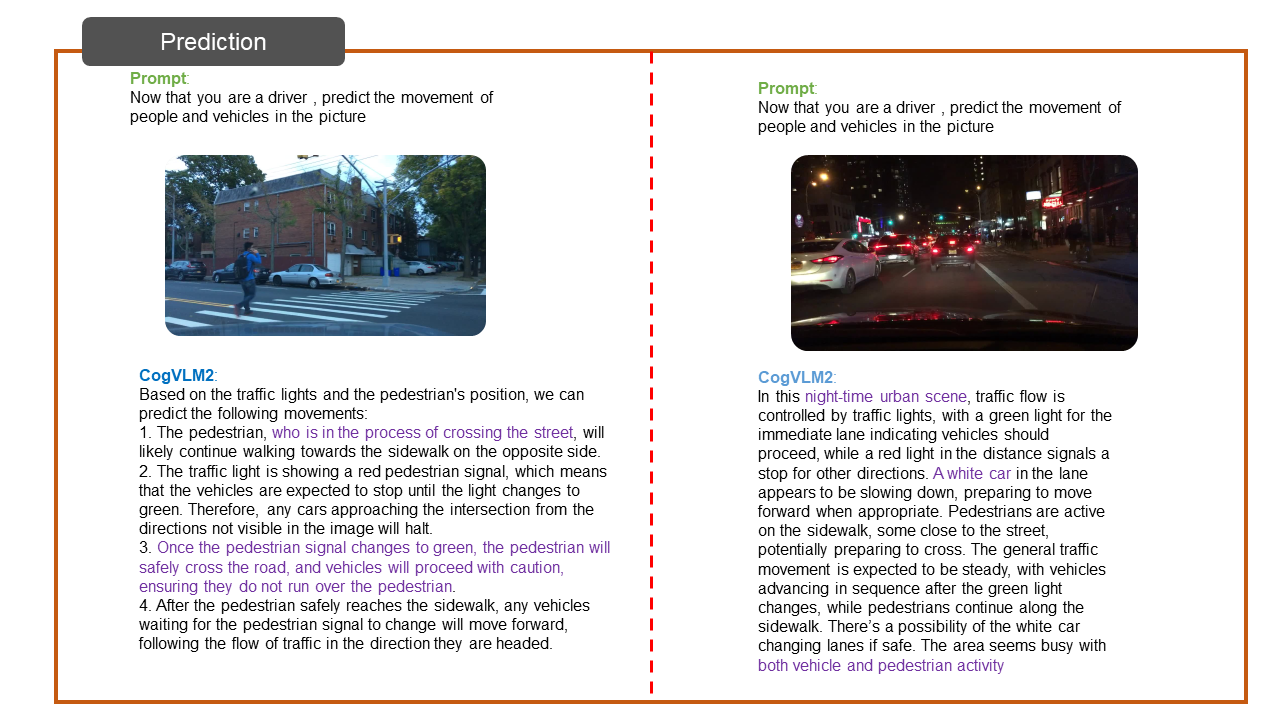}
    \caption{Prediction of traffic environments using fine-tuned CogVLM2. Correctly understood responses are indicated by purple font. Although the result shows
    a great ability to comprehend and describe visual stimuli, it is prone to occasional errors in judgment because of influences from unusual environmental
    contexts}
    \label{fig:prediction}
\end{figure*}

We tested our model's predictive skills by selecting scenarios with branched roads and decreased eyesight owing to rain, as shown in  Fig. \ref{fig:prediction} . The testing used a fine-tuned version of CogVLM2. The model effectively integrated and interpreted environmental cues for accurate prediction across all scenarios, demonstrating strong logical reasoning skills. The fine-tuned model can identify that one person is crossing the road. Also, it can identify the traffic light.

\subsection{Decision based on Multi-modal LLM}
\begin{figure*}[!h]
    \centering
    \includegraphics[width=\linewidth]{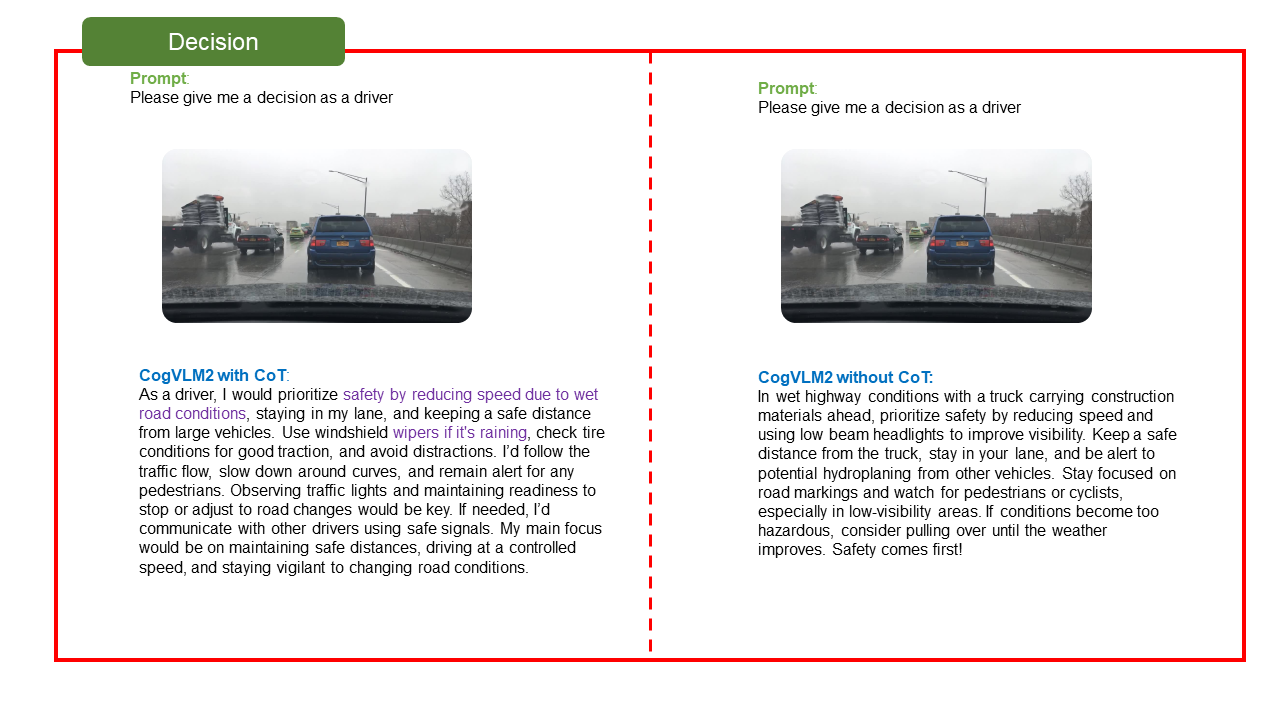}
    \caption{The left image is showing decision with Chain of Thought(CoT) and the right one is without CoT}
    \label{fig:decision}
\end{figure*}

This investigation examined the impact of CoT on decision-making by comparing CogVLM2 outcomes with and without CoT. The model in Fig. \ref{fig:decision} left panel was guided by the thought process outlined in Sec III-B. This approach helped the model integrate and interpret input from the thought chain, precisely detecting the vehicle's lane and make appropriate driving decisions. CoT plays a crucial influence in the model's ability to handle complex information. The fine-tuned CogVLM2 can identify the road is wet due to rain to the speed should be reduced also it suggests to use wippers.

\subsection{Simulation}
\begin{figure*}[!h]
    \centering
    \begin{tabular}{c@{\hspace{1em}}c@{\hspace{1em}}c@{\hspace{1em}}c}
        & \textbf{Ours} & {RL} & {MPC} \\ 
        \rotatebox{90}{{Roundabout}} &
        \includegraphics[width=0.28\textwidth]{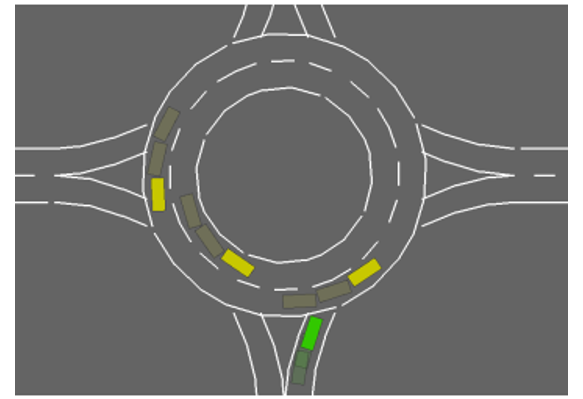} &
        \includegraphics[width=0.28\textwidth]{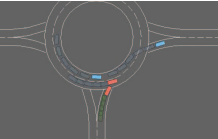} &
        \includegraphics[width=0.28\textwidth]{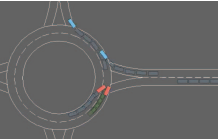} \\ 

        \rotatebox{90}{{Intersection}} &
        \includegraphics[width=0.28\textwidth]{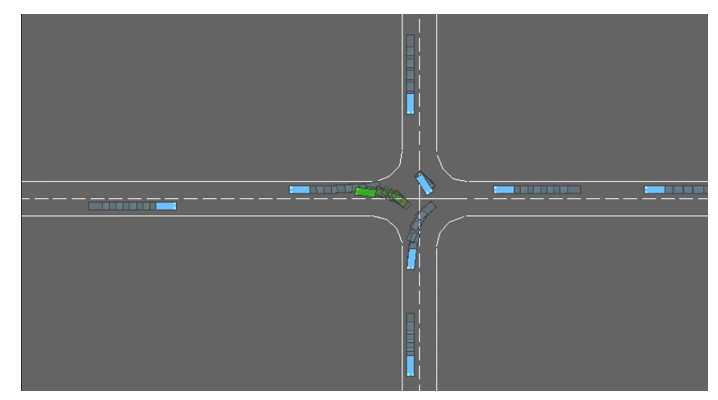} &
        \includegraphics[width=0.28\textwidth]{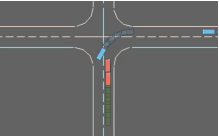} &
        \includegraphics[width=0.28\textwidth]{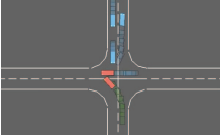} \\ 

        \rotatebox{90}{{Highway}} &
        \includegraphics[width=0.28\textwidth]{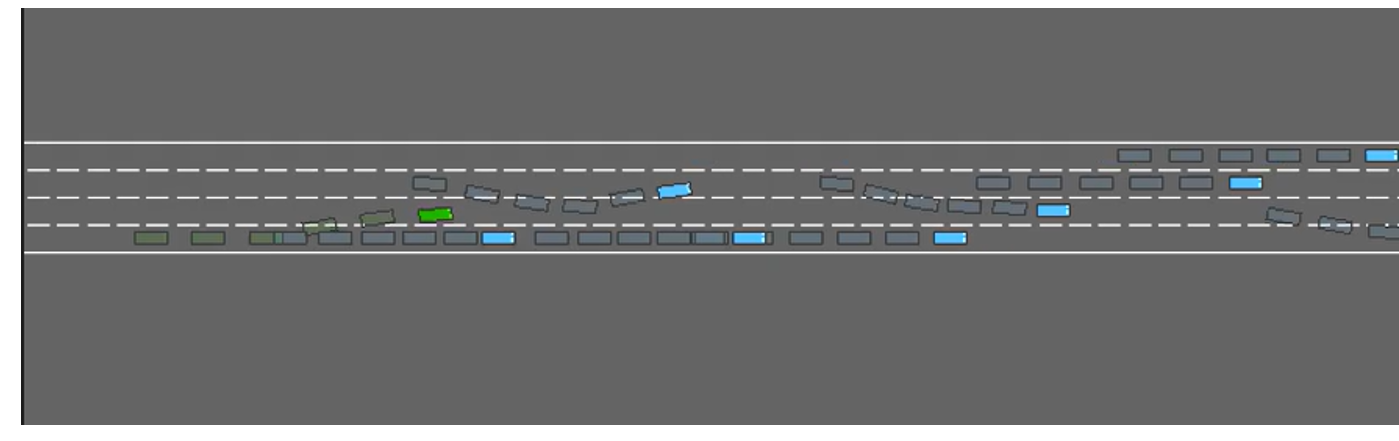} &
        \includegraphics[width=0.28\textwidth]{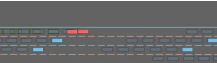} &
        \includegraphics[width=0.28\textwidth]{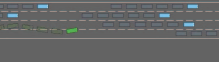} \\ 
    \end{tabular}
    \caption{We compared our technique against RL and MPC in three different scenarios: roundabout, intersection, and highway. In many cases, RL, motivated by time incentives, resulted in excessively aggressive driving and rear-end crashes. MPC, on the other hand, was unable to foresee numerous unexpected scenarios, resulting in task failures..}
    \label{fig:grid}
\end{figure*}

\begin{figure*}
    \centering
    \includegraphics[width=\linewidth]{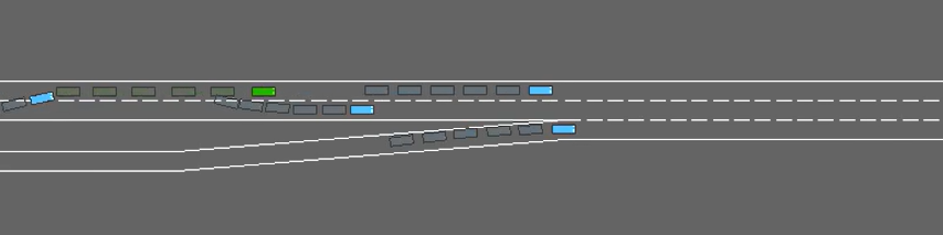}
    \caption{Road merging}
    \label{fig:merge}
\end{figure*}

Our suggested method was compared to RL and MPC in the Highway-env simulation environment [23], which included scenarios such as highways, roundabouts, and intersections. We created sophisticated real-world scenarios to evaluate the system's overall capabilities. We tested 50 different traffic scenarios across three environments. Table I displays the findings of measuring failure probability, ineffective decision likelihood, and average job completion time. In Fig \ref{fig:grid}, we present classic situations and compare our method to previous methods. The MLLM-based autonomous driving method makes better predictions about the surrounding environment. In the intersection situation, our technique accurately forecasted oncoming traffic and accelerated accordingly to merge into the lane. In the highway situation, our technique preemptively selected a lane with no vehicles for lane changing.
In contrast, RL, motivated by time constraints, caused collisions while rushing to achieve the destination. MPC cannot generalize rare events, such as rapid lane changes in roundabouts or the front car slowing down to yield in intersections, which can lead to crashes. Road margin is difficult for any vehicle, we also consider road merging, in Fig. \ref{fig:merge} we can see the ego vehicle can easily merge with another vehicle within a merging area.

\begin{table}[ht!]
  \centering
  \caption{The table displays measures where lower numbers indicate better performance. "Fail" denotes the percentage of job execution failures, whereas "Inefficiency" indicates the percentage of ineffective actions, and "Average Time" refers to the time required to accomplish a task at standard decision-making speeds, excluding decision-making time.
}
  \label{tab:performance_comparison}
  \begin{tabular}{@{}llccc@{}}
    \toprule
    \textbf{Scenario} & \textbf{Method} & \textbf{Fail} & \textbf{Inefficiency} & \textbf{Average Time} \\ 
    \midrule
    {Intersection} & RL & 10.0\% & 6.0\% & 3.9s \\ 
                                   & MPC & 4.0\% & 6.0\% & 4.0s \\ 
                                   & \cite{chen2024advanced} & 0.0\% & 2.0\% & 3.8s \\ 
                                   & \textbf{Ours} & 0.0\% & 3.0\% & 3.1s \\ 
    \midrule
    {Roundabout}   & RL & 10.0\% & 8.0\% & 4.8s \\ 
                                   & MPC & 6.0\% & 4.0\% & 5.2s \\ 
                                   & \cite{chen2024advanced} & 0.0\% & 2.0\% & 5.1s \\ 
                                   & \textbf{Ours} & 0.0\% & 2.3\% & 4.6s \\ 
    \midrule
    {Highway}      & RL & 12.0\% & 8.0\% & 18.2s \\ 
                                   & MPC & 6.0\% & 4.0\% & 19.1s \\ 
                                       & \cite{chen2024advanced} & 0.0\% & 2.0\% & 22.3s \\ 
                                   & \textbf{Ours} & 0.0\% & 5.0\% & 23.6s \\ 
                                       \midrule
    {Merge}    &  \textbf{Ours} & 3.0\% & 4.7\% & 14.9s \\ 
    \bottomrule
  \end{tabular}
\end{table}

\section{Discussion}

\subsection{Capabilities of Multi-modal Autonomous Driving}
\textbf{Scene Understanding}:
MLLMs can effectively interpret road scenes using inputs from cameras and sensors.
They identify key elements in the environment, such as:
Road Types: Dual and single carriageways.
Traffic Conditions: Real-time assessment of congestion or traffic flow.
Objects: Accurate recognition of vehicles, pedestrians, and other road users. \newline
\textbf{Contextual Decision-Making}:
By integrating multimodal data, MLLMs provide contextual insights to support decision-making.
They can evaluate scenarios like abnormal parking or unusual traffic patterns and suggest appropriate responses. \newline
\textbf{Handling Multimodal Inputs}:
MLLMs combine visual and textual data to provide comprehensive analysis and predictions.
For example, interpreting road images while understanding navigational instructions. \newline
\textbf{Strengths in Standard Driving Conditions}:
MLLMs demonstrate high accuracy in:
Identifying vehicles and pedestrians.
Assessing road conditions in typical driving environments.
Navigating straightforward traffic scenarios

\subsection{Limitations of Multi-modal Autonomous Driving}

\textbf{Difficulty in Complex Scenarios}:
MLLMs struggle with highly dynamic and unpredictable traffic environments, such as:
Overcrowded roads with multiple agents (e.g., cars, pedestrians, motorcycles).
Unusual events like sudden lane changes or road obstructions.
In such scenarios, the model may misinterpret the scene, leading to incorrect decisions. \newline
\textbf{Errors in Recognizing Atypical Situations}:
MLLMs may misjudge abnormal situations, such as:
Incorrectly interpreting the orientation of vehicles due to unusual parking positions (e.g., a red sedan parked abnormally). \newline
Perceiving non-existent objects in complex traffic, creating "illusions" that impact decision-making. \newline
\textbf{Limited Generalization}:
Despite training on large datasets, MLLMs may fail to generalize effectively to unseen conditions, such as:
Rare or region-specific traffic patterns.
Weather variations (e.g., fog, heavy rain) that obscure visual inputs. \newline
\textbf{Dependency on High-Quality Multimodal Inputs}:
MLLMs rely heavily on accurate, synchronized inputs from cameras, LiDAR, and other sensors.
Any degradation in input quality (e.g., sensor failure, poor lighting) can significantly impact performance. \newline
\textbf{Computational Complexity}:
MLLMs require substantial computational resources for real-time processing in autonomous vehicles, which can limit their scalability and deployment in cost-sensitive systems. \newline
\textbf{Lack of Robust Ethical and Contextual Understanding}:
MLLMs lack nuanced ethical judgment, such as:
Prioritizing safety in morally ambiguous situations (e.g., unavoidable collisions).
Understanding local driving rules and societal norms. \newline
\textbf{Vulnerability to Adversarial Attacks}:
MLLMs are susceptible to adversarial inputs, where manipulated data (e.g., adversarial images or spoofed signals) can mislead the model. \newline
\textbf{Language Limitations}:  
All the images used in our fine-tuned dataset contain English instructions rather than instructions in other languages, such as Bangla \cite{nextword}.
\section{Conclusion}
This study highlights the potential of Multi-Modal Language Models (MLLMs) to serve as effective decision-making agents in autonomous driving (AD) systems. We introduced an innovative framework designed to operate on systems with computational capabilities similar to those found in autonomous vehicles. Through extensive experiments conducted in multimodal, few-shot, and complex scenarios, we focused on evaluating the performance of MLLMs in scene perception, prediction, and decision-making. The results clearly demonstrated the significant advantages of integrating MLLMs into AD systems.

Additionally, this study examines the strengths and limitations of existing methodologies and proposes targeted strategies for future development. Our approach represents an initial step toward the development of MLLM-driven AD systems that prioritize safety, support few-shot learning, enable local deployment, and offer interpretability. We hope this work inspires further research and innovation in this promising field.

\bibliographystyle{IEEEtran}
\bibliography{bibliography}

\end{document}